\documentclass{article}

\usepackage{amsmath}
\usepackage{arxiv}

\usepackage[utf8]{inputenc} 
\usepackage[T1]{fontenc}    
\usepackage{hyperref}       
\usepackage{url}            
\usepackage{booktabs}       
\usepackage{amsfonts}       
\usepackage{nicefrac}       
\usepackage{microtype}      
\usepackage{lipsum}
\usepackage{graphicx} 

\title{Efficient Document Retrieval with G-Retriever }

\author{
  Manthankumar Solanki \\
  University of Stuttgart, Germany\\
  \texttt{st191474@stud.uni-stuttgart.de} \\
}

\begin{document}
\maketitle

\begin{abstract}
Textual data question answering has gained significant attention due to its growing applicability. Recently, a novel approach leveraging the Retrieval-Augmented Generation (RAG) method was introduced, utilizing the Prize-Collecting Steiner Tree (PCST) optimization for sub-graph construction. However, this method focused solely on node attributes, leading to incomplete contextual understanding. In this paper, we propose an enhanced approach that replaces the PCST method with an attention-based sub-graph construction technique, enabling more efficient and context-aware retrieval. Additionally, we encode both node and edge attributes, leading to richer graph representations. Our method also incorporates an improved projection layer and multi-head attention pooling for better alignment with Large Language Models (LLMs). Experimental evaluations on the WebQSP dataset demonstrate that our approach is competitive and achieves marginally better results compared to the original method, underscoring its potential for more accurate question answering.
\end{abstract}
\noindent Code available at: \href{https://github.com/manthan2305/Efficient-G-Retriever}{https://github.com/manthan2305/Efficient-G-Retriever}

\section{Introduction}
Large Language Models (LLMs) have significantly advanced natural language processing, particularly in question answering and conversational AI. Their integration with graph-structured data is gaining attention due to the prevalence of graphs in real-world applications, including knowledge graphs and social networks. Recent approaches combine LLMs with graph neural networks (GNNs) \cite{scarselli2009graph} to enhance reasoning over complex graphs \cite{jin2023llm, pan2023integrating}. However, efficiently leveraging LLMs for question answering on large textual graphs remains challenging, particularly in sub-graph construction and context-aware encoding.  

G-Retriever \cite{he2024gretriever} introduced a Retrieval-Augmented Generation (RAG) framework for question answering over textual graphs. It utilized the Prize-Collecting Steiner Tree (PCST) optimization for sub-graph construction, focusing solely on node attributes. Although effective, this method was limited by its exclusion of edge attributes and the complexity of PCST.  

To address these limitations, we propose an attention-based sub-graph construction technique that replaces PCST, enhancing retrieval efficiency and context-awareness. Our approach also encodes both node and edge attributes, leading to richer graph representations. Additionally, we incorporate an improved projection layer and multi-head attention pooling for optimized information aggregation.  

Experiments on the WebQSP dataset \cite{yih2016value} show that our method is competitive and achieves marginally better results than G-Retriever. These findings highlight the effectiveness of our approach in enhancing sub-graph construction and context-aware question answering.  

The main contributions of this paper are as follows:
\begin{itemize}
    \item We introduce an attention-based sub-graph construction method, replacing PCST for more efficient retrieval.
    \item Our approach encodes both node and edge attributes, leading to richer graph representations.
    \item We utilize an improved projection layer and multi-head attention pooling, achieving marginally better performance.
    \item Experimental results on the WebQSP dataset demonstrate the competitive performance of our method.
\end{itemize}

\section{Related Work}
\label{sec:headings}

\subsection{Graphs and Large Language Models (LLMs)}
Graphs serve as a fundamental structure for representing real-world relational data, making the integration of graph neural networks (GNNs) and large language models (LLMs) increasingly important for processing structured information \cite{he2024gretriever}. Research in this area spans graph reasoning\cite{chai2023graphllm}, node and graph classification \cite{yu2023leveraging,zhao2023gimlet}, multi-modal architectures \cite{yoon2023multimodal}, and LLM-driven knowledge graph tasks \cite{tian2023graph}, further demonstrating the potential of graph-enhanced language models in structured data interpretation.

GNNs enhance LLMs by providing graph-based reasoning capabilities, particularly in domains where textual attributes are embedded within structured relationships \cite{li2023survey}. This integration facilitates conversational interactions with graph data, enabling users to pose queries and receive contextually relevant responses grounded in structured knowledge. G-Retriever exemplifies this approach by combining graph retrieval and LLM-based reasoning, allowing for intuitive graph exploration. 

\subsection{Retrieval-Augmented Generation}
Retrieval-Augmented Generation (RAG) \cite{lewis2020retrieval} enhances large language models (LLMs) by retrieving relevant external knowledge before generating responses, making it particularly effective for tasks involving large knowledge graphs and structured data sources. Instead of relying solely on parametric memory, RAG-based models incorporate retrieved context, improving their ability to answer factual and multi-hop reasoning queries with higher accuracy. Additionally, RAG mitigates hallucination in LLMs by grounding their responses in retrieved factual knowledge, thereby increasing trustworthiness and explainability \cite{gao2023retrieval}.

\begin{figure}[h]
    \centering
    \includegraphics[width=0.8\linewidth]{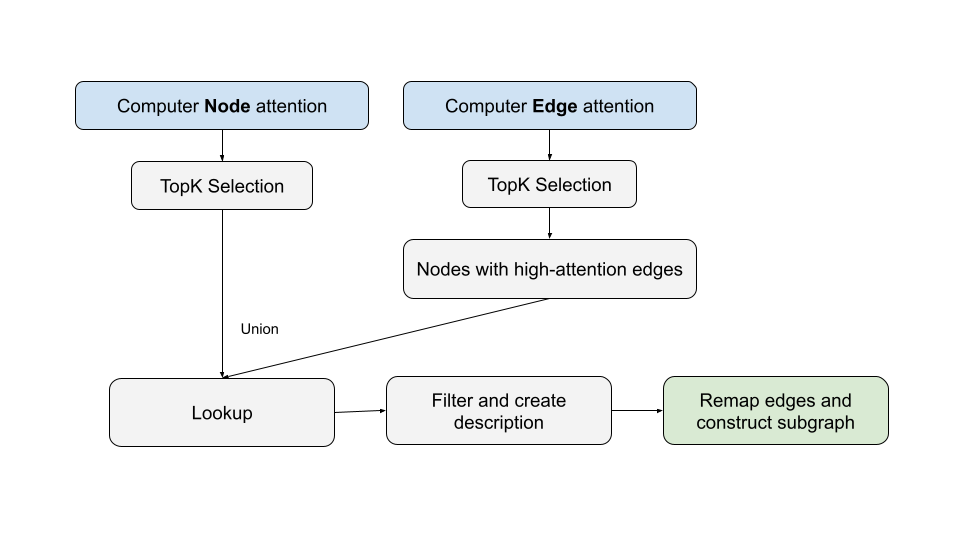}
    \caption{Retrieval via attention}
    \label{fig:rva}
\end{figure}

G-Retriever applies RAG principles to graph-based retrieval, where it retrieves query-relevant subgraphs. To achieve this, G-Retriever employs the Prize-Collecting Steiner Tree (PCST) \cite{bienstock1993prize} optimization method , which constructs a minimal subgraph connecting query-relevant entities while balancing node rewards and edge costs. Our research work includes replacement of PCST with attention driven subgraph selection (Figure ~\ref{fig:rva}), ensuring the retrieval of the most relevant entities and relationships.

\subsection{Attention mechanism}
Attention mechanisms significantly enhance the performance of graph neural networks (GNNs) and large language models (LLMs) by dynamically prioritizing relevant entities and relationships in graph-structured data \cite{knyazev2019understanding}. They filter noise by focusing on task-relevant components, improving retrieval precision and reasoning. Graph Attention Networks (GATs) \cite{velickovic2017graph} compute edge importance through learned attention scores, ensuring that retrieval is tailored to each query.

We propose multi-head attention pooling (MHA-POOL) to refine representation learning by assigning varying attention scores to node embeddings, enabling context-aware retrieval. Additionally, attention mechanisms improve graph pruning by filtering irrelevant entities and refining subgraph selection. By integrating these techniques in the G-Retriever we enhance the performance in knowledge-based question answering.

\section{Method}
Building upon the framework introduced by G-Retriever \cite{he2024gretriever}, our method introduces several key enhancements aimed at improving sub-graph construction efficiency and contextual understanding in question answering over textual graphs. Specifically, we propose two major amendments:

\subsection{Retrieval via Attention Mechanism}

We introduce an attention-based retrieval mechanism, replacing the Prize-Collecting Steiner Tree (PCST) method used in G-Retriever \cite{he2024gretriever}. This approach computes attention scores using cosine similarity between the query embedding and graph features, allowing dynamic selection of the most relevant nodes and edges. 

\textbf{Node Selection}: 
   - Compute node attention scores using cosine similarity between the query embedding \( q_{\text{emb}} \) and node features \( x \):
     \[
     \text{node\_scores} = \text{cosine\_similarity}(q_{\text{emb}}, x).
     \]
   - Select the top \( k \) nodes based on the highest scores or a threshold:
     \[
     V_{\text{topk}} = \{ v_i \mid \text{node\_scores}(v_i) \geq \text{threshold\_node} \} \cup \{ \text{top } k \text{ nodes} \}.
     \]

\textbf{Edge Selection}:
   - Compute edge attention scores using cosine similarity between the query embedding \( q_{\text{emb}} \) and edge features \( \text{edge\_attr} \):
     \[
     \text{edge\_scores} = \text{cosine\_similarity}(q_{\text{emb}}, \text{edge\_attr}).
     \]
   - Select the top \( k \) edges based on the highest scores or a threshold:
     \[
     E_{\text{topk}} = \{ e_{ij} \mid \text{edge\_scores}(e_{ij}) \geq \text{threshold\_edge} \} \cup \{ \text{top } k \text{ edges} \}.
     \]

\textbf{Subgraph Construction}:
   - Aggregate nodes that are incident to the selected edges:
     \[
     V_{\text{incident}} = \{ v_i \mid v_i \in \text{endpoints}(E_{\text{topk}}) \}.
     \]
   - Take the union of the top nodes and incident nodes:
     \[
     V^* = V_{\text{topk}} \cup V_{\text{incident}}.
     \]
   - Filter edges to ensure both endpoints are in \( V^* \):
     \[
     E^* = \{ e_{ij} \mid v_i, v_j \in V^* \}.
     \]

\textbf{Final Subgraph}:
   - The final subgraph \( S^* \) is defined as:
     \[
     S^* = (V^*, E^*),
     \]
   where \( V^* \) is the set of selected nodes and \( E^* \) is the set of selected edges.

Here, \( \text{node\_scores} \) and \( \text{edge\_scores} \) are computed using cosine similarity, and the subgraph \( S^* \) is constructed by ensuring that all selected edges are incident to the selected nodes. This approach ensures that the subgraph is both relevant to the query and efficiently constructed.

\subsection{Enhanced Graph Encoder}

The Enhanced Graph Encoder is designed to effectively encode both node and edge attributes, preserve the encoded information through multi-head attention pooling, and align the output with a large language model (LLM) by enhancing the projection layer. Below, we describe each component in detail.

\subsubsection{Joint Node-Edge Encoding:}  
To enrich graph representations, we jointly encode both node and edge attributes, effectively increasing the feature size. Edge features are first processed through a dedicated feed-forward network and then augmented with relative positional encodings derived from the differences between connected node features.

These enhanced edge representations are integrated with node features via Transformer convolution layers with residual connections, resulting in context-aware embeddings that capture richer structural information.

\medskip

\subsubsection{Multi-head Attention Pooling:}  
To derive a global representation from the node embeddings, we employ a multi-head attention pooling mechanism. In this module, multiple attention heads compute scalar attention scores for each node. These scores are normalized across nodes, and a weighted sum of the node embeddings is computed for each head. The outputs of all heads are then concatenated to form a comprehensive graph-level embedding that encapsulates diverse semantic and structural cues from the graph.
The updated representation is given by:
\begin{equation}
    h_g = \text{MHA-POOL} \left( \text{GNN}_{\phi_1} (S^*) \right) \in \mathbb{R}^{d_g}
\end{equation}

where:
\begin{itemize}
    \item \( S^* = (V^*, E^*) \) represents the retrieved subgraph, where \( V^* \) and \( E^* \) denote the set of nodes and edges, respectively.
    \item \( \text{GNN}_{\phi_1} (S^*) \) denotes the node embeddings obtained from the encoder.
    \item \( d_g \) represents the encoder output dimension.
\end{itemize}

\medskip

\subsubsection{Enhanced Projection Layer:}  
To align the enriched graph representations with the input space of the Large Language Model (LLM), we enhance the projection layer. This module is implemented as a two-layer MLP that first expands the feature dimensionality before projecting it back to the required size. By incorporating Layer Normalization and additional parameters, the projection layer better preserves edge information during the transformation. This enhanced alignment bridges the gap between the graph encoder and the LLM, thereby improving downstream performance in question answering tasks.

\section{Experiments}

\subsection{Reproducible Results}
We evaluate our approach using three model configurations on the WebQSP dataset:
\begin{enumerate}
    \item \textbf{Inference-only}: This configuration uses a frozen Large Language Model (LLM) for direct question answering without any fine-tuning or prompt adaptation. It serves as a baseline for zero-shot performance.
    
    \item \textbf{Frozen LLM with Prompt Tuning (PT)}: In this setup, the parameters of the LLM remain frozen, and only the prompt is adapted to improve performance. This allows the model to leverage task-specific information without modifying its core parameters.
    
    \item \textbf{Tuned LLM}: Here, the LLM is fine-tuned using Low-Rank Adaptation (LoRA), which introduces a small number of trainable parameters to adapt the model to the task. This configuration aims to balance performance and computational efficiency.
\end{enumerate}

We compare our results with those reported in the paper (denoted as "There") and our reproduced results (denoted as "Our"). The results are summarized in Table~\ref{tab:results}.

\begin{table}[h!]
\centering
\caption{Reproducible Results on WebQSP Dataset (Seed 0)}
\label{tab:results}
\begin{tabular}{|l|c|c|}
\hline
\textbf{Configuration} & \textbf{There} & \textbf{Our} \\ \hline
Inference-only - zero shot (Question only) & 41.06 & 42.99 \\ \hline
Frozen LLM w/ PT - Prompt Tuning & 48.34 ($\pm$ 0.64) & 52.94 \\ \hline
Frozen LLM w/ PT - G-Retriever & 70.49 ($\pm$ 1.21) & 72.72 \\ \hline
Tuned LLM - LoRA & 66.03 ($\pm$ 0.47) & 65.78 \\ \hline
Tuned LLM - G-Retriever w/ LoRA & 73.79 ($\pm$ 0.70) & 72.85 \\ \hline
\end{tabular}
\end{table}

Overall, our reproduced results align closely with the reported results, validating the robustness of the proposed configurations.

\subsection{Experiments with Model Architecture}
We conducted several experiments to improve the performance of the G-Retriever with LoRA configuration. The results of these experiments are summarized in Table~\ref{tab:model_experiments}.

\begin{table}[h!]
\centering
\caption{Experiments with Model Architecture on WebQSP Dataset}
\label{tab:model_experiments}
\begin{tabular}{|l|c|}
\hline
\textbf{Experiment} & \textbf{Test Accuracy} \\ \hline
Paper Results & 73.79 ($\pm$ 0.70) \\ \hline
Reproduced Results & 72.85 \\ \hline
Projection and Graph Encoder ((basic changes)) & 71.68 \\ \hline
Multi-Head Attention, Projection (more parameters) and Graph Encoder (improved) & 73.64 \\ \hline
\textbf{Subgraph Construction via Attention (Paper model)} & \textbf{74.14} \\ \hline
\textbf{Combined Enhancements (last two)} & \textbf{74.20} \\ \hline
\end{tabular}
\end{table}

\noindent
\textbf{Analysis of Results:}
The combined enhancements, which include improved projection layers, multi-head attention pooling, and subgraph construction via self-attention, achieve the highest accuracy of \textbf{74.20}, outperforming all previous configurations.

\section{Conclusion}
In this paper, we introduced an enhanced approach for question answering over textual graphs by replacing the PCST method with an attention-based sub-graph construction technique and encoding both node and edge attributes. Our method, incorporating an improved projection layer and multi-head attention pooling, achieved marginally better results on the WebQSP dataset compared to the original G-Retriever framework. These improvements demonstrate the effectiveness of our approach in enhancing context-aware retrieval and graph representation, paving the way for more accurate and scalable question-answering systems.

Future work will focus on further refining sub-graph generation methods and developing more robust graph encoding techniques to improve scalability and accuracy.

\bibliographystyle{unsrt}  
\bibliography{references}  

\section{Appendix}

\subsection{Training and Validation Loss}

In this section, we present the training and validation loss curves to demonstrate the effectiveness of our method in comparison to the original method. 

\begin{figure}[h]
    \centering
    \includegraphics[width=0.8\linewidth]{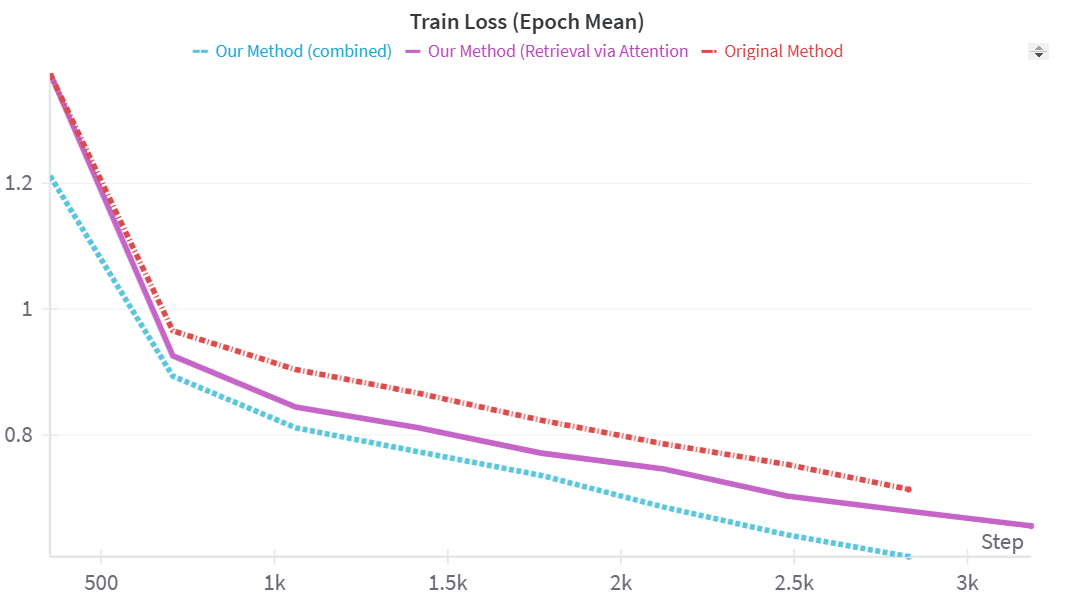}
    \caption{Training loss comparison between \textit{Our Method (combined)}, \textit{Our Method (Retrieval via Attention)}, and the \textit{Original Method}. Our proposed approach consistently shows lower training loss, indicating better optimization and convergence.}
    \label{fig:train_loss}
\end{figure}

\begin{figure}[h]
    \centering
    \includegraphics[width=0.8\linewidth]{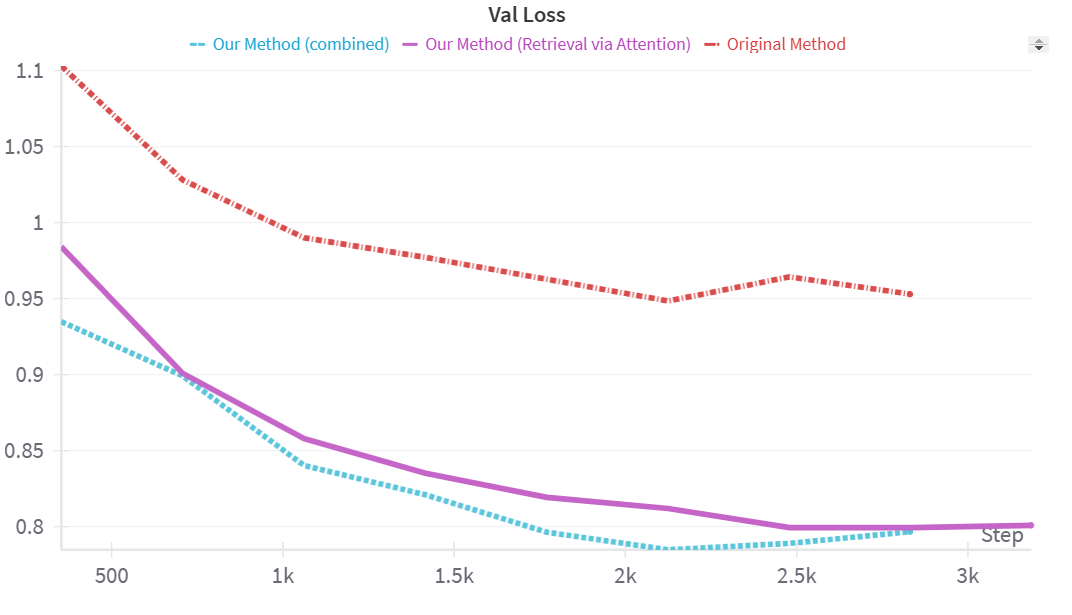}
    \caption{Validation loss comparison among different methods. Our proposed method exhibits a lower validation loss, confirming improved generalization performance.}
    \label{fig:val_loss}
\end{figure}

The results in Figures~\ref{fig:train_loss} and~\ref{fig:val_loss} indicate that \textit{Our Method (combined)}, which integrates Graph Encoder improvements with Retrieval via Attention, consistently outperforms the \textit{Original Method} across both training and validation phases. Additionally, \textit{Our Method (Retrieval via Attention)} also demonstrates improvements over the \textit{Original Method}, verifying the benefits of our retrieval strategy even when using the original architecture.

These findings suggest that our approach not only optimizes the training process more efficiently but also enhances the model’s ability to generalize to unseen data, as evidenced by the consistently lower losses across training, validation, and testing.

\end{document}